\documentclass{Interspeech}

\interspeechcameraready 

\title{Automatically assessing oral narratives of Afrikaans and isiXhosa children}

\author[affiliation={1}]{Retief}{Louw}
\author[affiliation={1}]{Emma}{Sharratt}
\author[affiliation={1}]{Febe}{de Wet}
\author[affiliation={1}]{Christiaan}{Jacobs}
\author[affiliation={2}]{Annelien}{Smith}
\author[affiliation={1}]{Herman}{Kamper}

\affiliation{Electrical and Electronic Engineering}{Stellenbosch University}{South Africa}
\affiliation{Speech, Language and Hearing Therapy}{Stellenbosch University}{South Africa}

\email{\{retieflouw,emsharratt\}@gmail.com}

\usepackage[prependcaption,textsize=scriptsize]{todonotes}

\usepackage{comment}
\definecolor{Ecolor}{HTML}{74a800}

\usepackage{cite}
\usepackage{balance}

\usepackage{booktabs}
\usepackage{multirow}
\usepackage{tabularx}
\newcommand{\mytable}{
	\centering
	\renewcommand{\arraystretch}{1.2}
}
\newcolumntype{C}{>{\centering\arraybackslash}X}
\newcolumntype{L}{>{\raggedright\arraybackslash}X}
\newcolumntype{R}{>{\raggedleft\arraybackslash}X}
\newcolumntype{P}[1]{>{\raggedright\arraybackslash}p{#1}}
\usepackage{siunitx}
\usepackage{etoolbox}                           
\newcommand{\ubold}{\fontseries{b}\selectfont}  
\robustify\ubold                                

\graphicspath{{fig/}}
\usepackage{pgfplots}
\usepackage{graphicx}
\usepackage[utf8]{inputenc}
\usepackage{tcolorbox}

\usepackage{microtype}

\usepackage[prependcaption,textsize=scriptsize]{todonotes}
\definecolor{mycolor}{HTML}{FF6600}
\definecolor{coolcolor}{HTML}{26209e}
\definecolor{plancolor}{HTML}{aaa6ed}
\definecolor{Ecolor}{HTML}{74a800}

\begin{document}

\maketitle

\begin{abstract}

Developing narrative and comprehension skills in early childhood is critical for later literacy.
However, teachers in large preschool classrooms struggle to accurately identify students who require intervention.
We present a system for automatically assessing oral narratives of preschool children in Afrikaans and isiXhosa.
The system uses automatic speech recognition followed by a machine learning scoring model to predict narrative and comprehension scores.
For scoring predicted transcripts, 
we compare a linear model to a large language model~(LLM). 
The LLM-based system outperforms the linear model in most 
cases, but the linear 
system is competitive despite its simplicity.
The LLM-based system is comparable to a human expert in flagging children who require intervention.
We lay the foundation for automatic oral assessments in classrooms, giving teachers extra capacity to focus on personalised support for children's 
learning.

\end{abstract}

\section{Introduction}
\label{sec:intro}
Oral storytelling and comprehension are essential skills for children, and they are strong predictors for later reading proficiency~\cite{hayward_schneider_etal2009, chiu2018, hjetland_brinchmann_etal2020, babayigit_roulstone_etal2021}.
Assessing early oral narrative ability is therefore crucial for understanding language development and identifying delayed literacy~\cite{dickinson_mccabe_etal2003, schick_melzi2010, reese_leyva_etal2010, oakhill_cain2012}.
However, in many parts of the world, large classroom sizes pose significant challenges for preschool educators.
In such contexts, observational assessments, which depend on teacher intuition, are often inaccurate~\cite{shermis_burstein2003, mozer_miratrix_etal2023}.
On the other hand, one-on-one assessments are effective for diagnosing students’ language skills~\cite{chiquito2023effects}.
However, the capacity to do individual assessments is often limited by the time, cost, and logistical challenges involved in scaling across large classrooms~\cite{snow_matthews2016}.

Our goal is to develop a system to support preschool teachers in low-resource language settings by identifying children whose oral language skills are below the level expected at their age.
The system should facilitate more consistent and efficient evaluations, even in classrooms with large numbers of learners who are too young for written assessments.
We focus on two language communities in South Africa, where literacy is a major problem. 
Only 20\% of 10-year-olds can read for meaning~\cite{roux_van_staden_etal2023} and overcrowded preschool classrooms are a norm.

\begin{figure}[!h]
    \centering
    \includegraphics[width=0.98\linewidth]{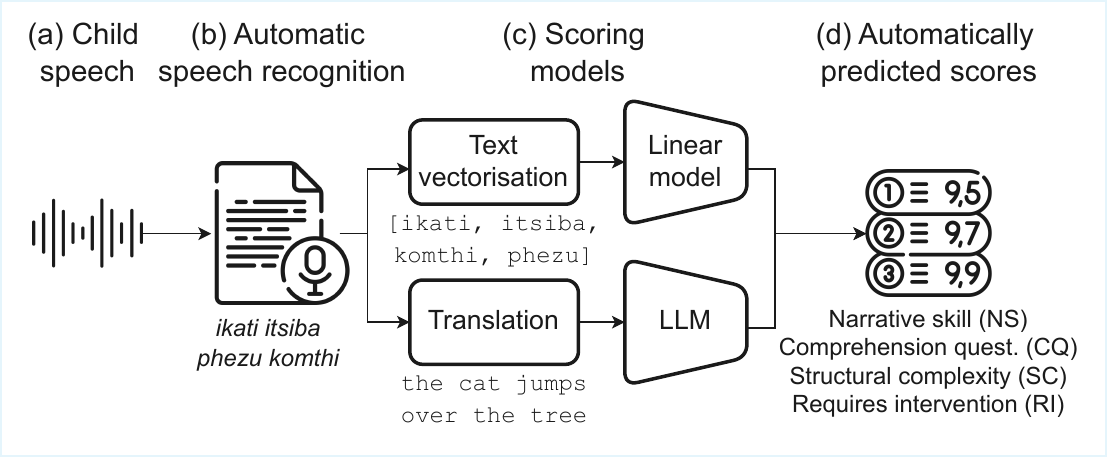}
    \caption{Our system for automatically scoring narratives told by children aged four to five. An isiXhosa example is shown.}
    \label{fig:system}
\end{figure}

We specifically present the development of an oral narrative assessment system for Afrikaans- and isiXhosa-speaking children between the ages of four and five.
A child is shown a series of pictures, prompting them to tell a story. 
Afterwards, they need to answer a series of comprehension questions.
As illustrated in Figure~\ref{fig:system}, the system takes the child's speech as input.
It should then predict as output a range of finer-grained scores, such as narrative cohesion and comprehension, as well as coarser scores, like a binary value for whether intervention is required.

We develop separate Afrikaans and isiXhosa systems. 
Our proposed methodology consists of an automatic speech recognition (ASR) model that transcribes the child’s speech, as shown in Figure~\ref{fig:system}(b).
The text output from the ASR model is then processed (c) by one of two scoring models: a linear model or a large language model (LLM).
For the linear method, simple features like word count and narrative length are extracted, serving as input to a regularised model that predicts the scores.
For the LLM-based approach, the ASR output is first automatically translated from Afrikaans or isiXhosa to English and then passed on to an LLM, which is prompted using in-context learning to (d) predict scores.

We find that, with a few exceptions, the LLM performs better than the linear model.
Even though the error rates of the ASR models are high, the LLM-based approach is able to identify children requiring intervention with an accuracy of more than 80\% for Afrikaans and 64\% for isiXhosa.
For the same task, a speech therapist achieves 80\% for Afrikaans and 70\% for isiXhosa in a re-assessment test.
These imperfect human scores highlight the challenging and subjective nature of the task.
We analyse the effects of ASR and automatic translation, and look at differences between Afrikaans and isiXhosa.

Taken together, our findings indicate that early childhood assessment is a complex task, one that even a human professional finds challenging. 
Nevertheless, we show the potential of an automatic assessment system to assist, complement, and enhance human assessments
in real, low-resource settings.

\section{Data}
\label{sec:data}

\begin{figure}[!t]
    \centering
    \includegraphics[width=0.99\linewidth]{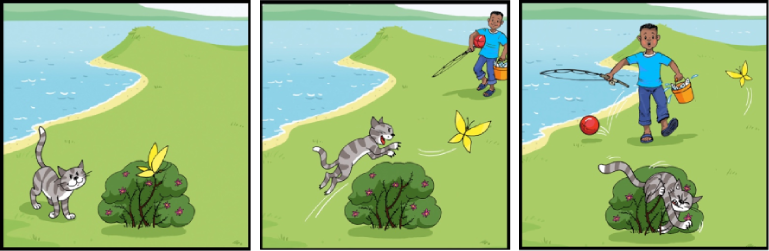}
    \caption{One of the picture sequences used in the MAIN assessment protocol to elicit a narrative from a child.
    }
    \label{fig:MAIN_kat_story}
\end{figure}

We develop our automatic assessment system using data from a study
that involved children aged four to five from lower socioeconomic backgrounds in South Africa~\cite{smith2023}.
Children were shown a series of six pictures and asked to tell a story in their home language of either Afrikaans or isiXhosa (Figure~\ref{fig:MAIN_kat_story}).
The children's speech was recorded using Samsung Galaxy Tab A7 Lite 8.7 tablets and Logitech H111 headsets.
The children were evaluated using the multilingual assessment instrument for 
narratives
(MAIN) protocol~\cite{gagarina_klop_etal2019}. 
This tool is standardised for South African populations, accounting for cultural biases.
Assessments were carried out by trained assessors experienced in early childhood development~\cite{smith2023}.

MAIN assesses three key areas. 
Narrative skill (NS), ranging from 0 to 16, measures narrative complexity, including inferred goal-directed behaviour (e.g., \textit{the cat wants to catch the butterfly}). 
Comprehension question (CQ) scores, from 0 to 10, are determined by responses to ten questions assessing understanding of story structure (e.g., \textit{why does the cat jump?}). 
NS and CQ prediction are regression tasks. 
Structural complexity (SC) categorises narratives into one of six classes, evaluating how effectively a child conveys goal-directed actions. 
Higher SC categories indicate complex narratives with goals, attempts, and outcomes (e.g., \textit{he wants to take his ball, he takes it, he plays with it}). 
SC prediction is a classification~task.
Children aged four should achieve at least an SC score within the third category~\cite{westby2005}, so the lowest two SC categories indicate developmental delays.
We assign a binary label to these cases that require intervention (RI), and measure performance on this binary classification task.

The data~\cite{smith2023} is split into train (roughly 200 children), development (38 children), and test (28 children) sets per language, with no speaker overlap. 
Each language has approximately 5 hours of active child speech, reflecting its low-resource nature.\footnote{Only the child's speech is included; the assessor's speech was manually removed. 
This is an idealised scenario, equivalent to using a perfect diarisation system. Future work will look at a complete system.}
Differences between Afrikaans and isiXhosa are immediately evident: isiXhosa is morphologically rich with an agglutinative orthography~\cite{mzamo_helberg_etal2015}, resulting in 3928 unique word types in our data, almost three times the 1140 types in the Afrikaans data.\footnote{isiXhosa and Afrikaans, with 8 million and 7.2 million speakers respectively, are two of South Africa's 12 official languages. 
isiXhosa is a Southern Bantu language, while Afrikaans is a West Germanic language derived from Dutch. 
Both languages use the Latin alphabet.}

\section{Method}
\label{sec:method}

Our automatic narrative assessment system uses an ASR model that feeds into a model that predicts scores, as shown in Figure~\ref{fig:system}.

\subsection{Automatic speech recognition (ASR) models}

For each of the languages, we use a corresponding ASR model.
These models were developed in previous work, with full details provided in~\cite{jacobs2024}.
In short, Whisper-based ASR models~\cite{radford2023whisper} 
are respectively trained on five minutes of transcribed child speech in Afrikaans and isiXhosa.
The training data is a subset of the training data described in Section~\ref{sec:data}.
We use this restricted subset for two reasons: (1) this is representative of extreme low-resource settings, and (2) it allows us to see the effect of ASR errors on the downstream score prediction task.
The character error rates~(CERs) for these models are 28\% for Afrikaans and 34\% for isiXhosa. 
isiXhosa exhibits a higher CER than Afrikaans, possibly because Whisper is pretrained on some adult Afrikaans and related languages like Dutch and Flemish, while isiXhosa is not included, nor any language closely related to isiXhosa.
We aim to determine the usefulness of these ASR models despite their high error rates.

\subsection{Linear model}
ASR output is fed into a subsequent scoring model, as shown in Figure~\ref{fig:system}(c). We consider two options. The first is a linear model. 
To predict NS and CQ, we use linear regression.
We employ logistic regression to predict the SC and RI categories.
Linear models are simple, can be efficiently trained with very little data, and are more interpretable than complex models~\cite{dawes1979}.

A range of features are used.
We derive speech duration features from the original recordings: duration, mean utterance length, and standard deviation.
Using the ASR transcripts, we obtain the total word token and type count as well as counts for individual word types, e.g., there is a feature counting how often \textit{ikati} (cat) occurs.
We experimented with lemmatising words, but this did not improve development performance.
Our final feature is the Flesch–Kincaid readability score, which quantifies linguistic complexity based on sentence length and syllable~count~\cite{flesch1948, kincaid1975}.
The score was originally designed for English, but can easily be used on other languages and gave benefits in our development experiments.

Other features were also considered in development experiments but did not improve score prediction performance: 
articulation rate, part-of-speech counts, type-token ratio, sentence complexity, and the Gunning Fog Index~\cite{kajzer2023into}.
We also explored morphological parsing for isiXhosa, replacing words with morphemes~\cite{moeng_reay_etal2021}, and using n-gram features in both languages, but this did not improve development scores. 

While linear models are well-suited for working with limited data, they remain susceptible to overfitting, especially when the number of input features exceeds the number of observations~\cite{ranstam_cook2018}, as is the case here. 
We therefore use lasso regularisation.
Lasso adds an L1 penalty $\alpha||\mathbf{w}||_1$ to a linear loss. 
This 
shrinks coefficients to 
zero, thereby selecting relevant features and excluding non-informative ones~\cite{james_witten_etal2023, jurafsky_martin2025}.
We tune $\alpha$
on development data. 
The result is a major reduction in the number of features. 
For instance, in the Afrikaans model for predicting NS, more than 1140~weights are zeroed leaving only 55~features.
We train distinct lasso-regularised linear 
models for predicting each assessment score in each language.

\subsection{Large language model (LLM)}
\label{sec:llms}

Our second approach uses an LLM with in-context learning to score the children's transcripts, as illustrated in Figure~\ref{fig:system}(c)-bottom.
Despite being multilingual and understanding Afrikaans and isiXhosa, LLMs are primarily trained on English text. 
To address this, we first translate the Afrikaans and isiXhosa ASR output to English using the Google Translate API. 
In Section~\ref{sec:results} we give results with and without translation.

In-context learning allows pre-trained LLMs to generate task-specific outputs by providing them with examples of ideal input-output pairs.
Literature about on in-context learning recommends several techniques of prompt engineering
to craft the ideal prompt~\cite{dong-etal-2024-survey}.
The prompt format shown in Figure~\ref{fig:prompt2} consistently yielded the best results and is used for all the final results.
The prompt includes training examples (each comprising a translation of the ASR output and the three corresponding teacher scores); the prompt instruction; and the unseen translated story. 

\begin{figure}[!t]
    \centering
    \begin{tcolorbox}[colframe=black!50, colback=black!3, sharp corners=all, boxrule=0.1mm, left=1mm, right=1mm, top=1mm, bottom=1mm]
    {\ttfamily\scriptsize
    \textbf{Examples from the training set}

    \begin{tcolorbox}[colframe=cyan!50, colback=cyan!10, sharp corners=all, boxsep=0.5pt, left=1mm, right=1mm, top=1mm, bottom=1mm, boxrule=0.1mm,arc=0pt,outer arc=0pt]
    Transcript: grass is here. the cat... \\
    \{NS: 4\}, \{CQ: 7\}, \{SC: 2\}
    \end{tcolorbox}

    \begin{tcolorbox}[colframe=cyan!50, colback=cyan!10, sharp corners=all, boxsep=0.5pt, left=1mm, right=1mm, top=1mm, bottom=1mm, boxrule=0.1mm,arc=0pt,outer arc=0pt]
    ...
    \end{tcolorbox}

    \textbf{Instruction}
    \begin{tcolorbox}[colframe=blue!50, colback=blue!10, sharp corners=all, boxsep=0.5pt, left=1mm, right=1mm, top=1mm, bottom=1mm, boxrule=0.1mm,arc=0pt,outer arc=0pt]
    Each of the above is a story followed with a score for NS, CQ and SC.
    NS and CQ are floating point numbers and SC is an ordinal category.
    Predict NS, CQ and SC for the following story. Only give the scores.
    \end{tcolorbox}

    \textbf{Unseen sample from the test set}

    \begin{tcolorbox}[colframe=teal!50, colback=teal!10, sharp corners=all, boxsep=1pt, left=1mm, right=1mm, top=1mm, bottom=1mm, boxrule=0.1mm,arc=0pt,outer arc=0pt]
    Transcript: the cat wants to kill the...
    \end{tcolorbox}
    }
    \end{tcolorbox}
    \vspace*{-5pt}
    \caption{The structure of how we prompt LLMs using in-context learning for scoring oral narratives from children.}
    \label{fig:prompt2}
\end{figure}

We conducted development experiments to identify the optimal in-context learning approach. 
We compared three language models: Gemini-1.5-Flash-8B~\cite{mike2025}, GPT-4~\cite{mcfayden2024chatgpt}, and Llama-3.3-70B~\cite{ibrahim2024fine}.
Gemini-1.5-Flash-8B consistently achieved the best development performance, so it was selected as the LLM for the system.
We also investigated several few-shot prompting strategies, such as target-based grouping and balanced sampling~\cite{zhou2023large}.
Performance consistently improved with more examples, so we use the entire training set in each prompt for the final system.
We also considered adding explanations of how scores are qualitatively determined by the teachers, but this did not improve development performance.

In a study similar to ours, Baumann et al.~\cite{baumann-etal-2024-bert} investigated automatic scoring of German children's MAIN narratives.
They fed a German BERT model with the text from human-transcribed narratives.
Aggregated embeddings were then given as input to linear models, predicting binary scores for each subcategory of NS.
Although they achieved usable results
for two NS subcategories, they encountered challenges with others.
Our study differs in several regards: we employ an LLM for direct scoring instead of training a classifier, use ASR outputs instead of transcriptions, focus on two low-resource languages instead of a high-resource language, and incorporate additional MAIN scores (CQ, SC) for holistic assessments.
Another related study is that of Veeramani et al.~\cite{veeramani23_slate}, who also used a BERT-based score prediction model, but fed it with actual ASR predictions.
However, they focused on narratives from  English children between the ages of nine and thirteen, while we consider younger children (between four and five) speaking low-resource languages.

\section{Experimental setup}
\label{sec:setup}

For assessing performance, we use \(R^2\) for the narrative (NS) and comprehension (CQ) score regression tasks, Cohen’s \(\kappa\) for structural complexity (SC) classification, and accuracy for identifying children requiring intervention (RI).

\(R^2\) quantifies a model's fit by measuring the explained variance in the target variable~\cite{Saunders_russell_etal2012, james_witten_hastie_etal2023}, with 0 for mean prediction and 1 for perfect predictions~\cite{Saunders_russell_etal2012}. 
Cohen’s \(\kappa\) quantifies agreement beyond chance~\cite{cohen1960} and assesses model alignment with ground truth in machine learning~\cite{grandini_bagli_etal2020}. 
Ranging from $-$1 to 1, \(\kappa\) indicates perfect agreement at 1 and chance-level at 0. 
Given that SC categories are ordinal (e.g., four is closer to five than to one), we use linear-weighted Cohen’s \(\kappa\), applying a linear penalty for diverging categories~\cite{nelson_edwards2015}.

To contextualise results, we compare them with human expert predictions: a speech therapist from the original study~\cite{smith2023} manually re-assessed ten samples per language.
The resulting \(R^2\), \(\kappa\), and accuracy scores establish human expert performance for
the complex task of child oral narrative assessment, 
which is inherently subjective~\cite{shepard1994}.

\section{Experimental results}
\label{sec:results}

Our goal is to determine the most effective automatic system for assessing narratives from preschool children.
Table~\ref{tbl:test_results_NS_CQ_SC} presents the test results for NS, CQ, SC, and RI, comparing the predictive performance of the linear models and LLMs. 
The \textit{oracle} columns show idealised results when ground truth transcriptions are used in training and evaluating the scoring models---these serve as a reference.
The
\textit{ASR} columns use automatically produced transcriptions.

\newcommand{\ManualLabel}{Oracle}
\begin{table}[!t]
    \caption{Test set performance of the automated narrative scoring system. We report $R^2$ for narrative skills (NS) and comprehension (CQ), Cohen's kappa ($\kappa$) for structural complexity (SC), and accuracy for children requiring intervention~(RI).}
    \label{tbl:test_results_NS_CQ_SC}
    \eightpt 
    \mytable
    \begin{tabularx}{\linewidth}{@{}l@{\ \ \ \ }CcCcCcCc@{}}
        \toprule
        & \multicolumn{2}{c}{NS [$R^2$]} & \multicolumn{2}{c}{CQ [$R^2$]} & \multicolumn{2}{c}{SC [$\kappa$]} & \multicolumn{2}{c}{RI [\%]} \\
        \cmidrule{2-3} \cmidrule(l){4-5} \cmidrule(l){6-7} \cmidrule(l){8-9}
        {} & \ManualLabel & ASR & \ManualLabel & ASR & \ManualLabel & ASR & \ManualLabel & ASR \\
        \midrule
        \multicolumn{2}{@{}l}{\underline{\textit{Afrikaans:}}} \\
        {Linear} & {0.27} & {0.20} & {0.28} & {0.03} & {0.18} & {0.12} & {68} & {61} \\
        {LLM} & {0.67} & \textbf{0.52} & {0.29} & \textbf{0.20} & {0.13} & \textbf{0.13} & {89} & \textbf{89} \\
        \addlinespace
        Human & {0.82} & {-} & {0.79} & {-} & {0.25} & {-} & {80} & {-} \\

        \addlinespace
        \multicolumn{2}{@{}l}{\underline{\textit{isiXhosa:}}} \\
        {Linear} & {0.49} & {0.32} & {0.38} & \textbf{0.10} & {0.01} & {0.19} & {43} & {54} \\
        {LLM} & {0.58} & \textbf{0.52} & {0.06} & {0.08} & {0.39} & \textbf{0.33} & {64} & \textbf{64} \\
        \addlinespace
        Human & \mbox{0.17} & {-} & 0.85 & - & 0.50 & - & {70} & {-} \\
        \bottomrule
    \end{tabularx}
\end{table}

\textbf{LLM vs linear.} 
We start by comparing the linear models to the LLMs 
for the systems using ASR (oracle results are discussed below).
Given its simplicity, the linear-based system performs competitively. 
But the LLM achieves higher scores in all cases, except for CQ on isiXhosa. 
As a concrete example, the LLM attains an $R^2$ of 0.52 when predicting NS in Afrikaans, whereas the linear model scores only 0.20.
On Afrikaans, the LLM outperforms the logistic regression model by almost 30\% absolute in predicting whether intervention is required (RI), while on isiXhosa, its RI accuracy is 10\% better.

Table~\ref{tbl:test_recall_F1_RA} reports additional recall and F1 scores for RI prediction, supporting our findings. 
We prioritise recall, as the system's goal is to identify children most in need. 
The LLM consistently outperforms the linear model across all cases in Table~\ref{tbl:test_recall_F1_RA}, effectively identifying children with learning difficulties. 
Table~\ref{tbl:test_recall_F1_RA} also presents the mean absolute error for NS (0 to 16) and CQ (0 to 10) predictions in the original units. 
The LLM consistently achieves lower error than the linear model for both NS and CQ. Overall, the LLM is superior for automatically identifying children requiring academic support.

\begin{table}[!t]
    \caption{Test set results in terms of mean absolute error (MAE) for the narrative skill (NS) and comprehension questions (CQ) scores, and recall and F1 (\%) for predicting whether children require intervention~(RI) using ASR outputs.}
    \eightpt
    \label{tbl:test_recall_F1_RA}
    \begin{tabularx}{\linewidth}{@{}l@{\ \ \ \ }CCCC@{}}
        \toprule
        & \multicolumn{2}{c}{MAE} & \multicolumn{2}{c}{RI} \\
        \cmidrule{2-3} \cmidrule(l){4-5}
        {} & NS & CQ & Recall & F1 \\
        \midrule
        \multicolumn{2}{@{}l}{\underline{\textit{Afrikaans:}}} \\
        {Linear} & {1.21} & {1.91} & {50} & {33} \\
        {LLM} & \textbf{0.90} & \textbf{1.45} & \textbf{96} & \textbf{94} \\
        \addlinespace
        {Human} & {0.50} & {0.90} & {89} & {89} \\
        \addlinespace
        \multicolumn{2}{@{}l}{\underline{\textit{isiXhosa:}}} \\
        {Linear} & {1.34} & {2.08} & {60} & {55} \\
        {LLM} & \textbf{0.96} & \textbf{1.64} & \textbf{75} & \textbf{78} \\
        \addlinespace
        {Human} & {0.80} & {0.38} & {100} & {77} \\
        \bottomrule
    \end{tabularx}
\end{table}

\textbf{Automatic system vs human.} 
We compare our system to a human expert from the original study~\cite{smith2023}, a speech and language therapist, who re-assessed a subset of the original data.
In Table~\ref{tbl:test_results_NS_CQ_SC}, the human achieves higher $R^2$ scores for NS in Afrikaans and outperforms the models in both languages when predicting SC and CQ.  
For isiXhosa RI, the LLM (64\%) approaches human performance (70\%). Interestingly, the LLM surpasses human scores in two instances: Afrikaans RI (89\% vs 80\%) and isiXhosa NS $R^2$ (0.52 vs 0.17).
We interpret these superior model scores with caution, as they reflect the limitations and subjectivity in the ground truth data.
Human assessment scores varied across all metrics, revealing potential biases and challenges in teacher-generated evaluations, emphasising the complexity of accurately assessing children.

\textbf{The effect of imperfect ASR.} 
To assess the impact of the ASR model's high CER (Section~\ref{sec:data}), we compare the oracle and ASR scores in Table~\ref{tbl:test_results_NS_CQ_SC}. 
Despite relatively high error rates, the expected drop in ASR scores is not substantial.
Notably, in some cases, ASR scores exceed oracle scores. 
For example, when classifying children that should be flagged, the isiXhosa linear model's score of 54\% is better than the oracle score of 43\%. 
These results show that even noisy ASR predictions can still be meaningful for downstream assessments.

\textbf{Afrikaans vs isiXhosa.}
We find that the LLM generally performs worse on RI for isiXhosa (64\%) compared to Afrikaans (89\%).
This discrepancy cannot be attributed solely to the isiXhosa ASR system's poorer CER, or the translation step, as the oracle RI results are also worse (64\% vs 89\%).
Future work may explore whether the lower isiXhosa performance stems from inherent properties of the 
data itself, or from limitations in the automatic English-to-isiXhosa translation process.
For most of the other metrics, results are similar between the two languages.

\textbf{The effect of translation.}
The LLM system only sees English text due to the intermediate translation step.
Table~\ref{tbl:results_transl} shows that translating the text from low-resource languages to English improves the LLM performance regardless of the language or score.
The improvement is particularly evident for Afrikaans children, where the $\kappa$ increases from 0.01 to 0.08 with translation.
The Google Translate API's apparent ability to handle noisy ASR predictions, including potentially changing words and fixing errors like stuttering, likely contributes to the improved performance.

We performed these translation experiments on the development set.
There is a substantial difference in \(R^2\) and \(\kappa\) between the final row of Table~\ref{tbl:results_transl} giving development-set scores and the LLM rows in Table~\ref{tbl:test_results_NS_CQ_SC} giving test-set scores.
For example, the LLM achieves an \(R^2\) of 0.41 for CQ on the development set, but only 0.20 on the test set.
This difference suggests that the specific children included in each set also have a considerable influence on the evaluations. 

\begin{table}[!t]
    \caption{
    Development set performance comparing LLMs with and without automatic translation to English before scoring.
    }
    \eightpt
    \label{tbl:results_transl}
    \mytable
    \begin{tabularx}{\linewidth}{@{}lCCCCCC@{}}
        \toprule
        & \multicolumn{3}{c}{Afrikaans} & \multicolumn{3}{c}{isiXhosa}  \\
        \cmidrule(l){2-4} \cmidrule(l){5-7}
        Model & NS [$R^2$] & CQ [$R^2$] & SC [$\kappa$] & NS       [$R^2$] & CQ [$R^2$] & SC [$\kappa$] \\
        \midrule
        LLM without transl. & {0.18} & {0.33} & {0.01} & {0.17} & {0.18} & {0.12}\\
        LLM with transl. & \textbf{0.23} & \textbf{0.41} & \textbf{0.08} & \textbf{0.30} & \textbf{0.19} & \textbf{0.13}\\
        \bottomrule
    \end{tabularx}
\end{table}

\textbf{Linear model interpretability.}
Although not always performing as well as the LLM, the linear model's weights can be inspected. 
Lasso regularisation pushes most weights to zero and we can examine the largest weights of the remaining features.
For predicting NS and CQ in Afrikaans, the largest weights are associated with the features counting words like
\textit{gelukkig} (happy), \textit{eet} (eat) and \textit{kwaad} (angry).
For isiXhosa, they include \textit{kabuhlungu} (painfully), \textit{ithathe} (take it) and \textit{ilambile} (hungry). 
For both languages, the largest weights are associated with
a combination of emotive, action-related and descriptive word types. 
In parallel work~\cite{sharratt_smith_etal2025},  we explicitly consider how the linear model can be used to gain deeper interpretable insights into the acoustic, lexical and semantic patterns of oral narratives from preschoolers requiring intervention.

In summary, our results show that an LLM-based system outperforms a system that uses a linear model to 
predict most assessment scores.
Giving results in the original units, for
predicting the NS score out of 16, our best system 
achieves a mean absolute error
of 0.90 for Afrikaans and 0.96 for isiXhosa, respectively.
While a human expert achieves higher scores in most metrics, the LLM shows comparable and even superior performance in the final intervention classification task, suggesting limitations in the human-generated ground truth data.  
Despite imperfect ASR transcripts, the system maintains performance, with translation further improving LLM results for both languages.

\section{Conclusion}
\label{sec:conclusion}

We introduced an automated 
system 
to assist early childhood educators in assessing children's oral narrative and comprehension skills in two low-resource languages. 
We specifically combined
automatic speech recognition (ASR) with two predictive models.
We developed linear models and prompted large language models (LLMs) to predict several development scores from stories told by Afrikaans and isiXhosa children. 
The LLM-based system aligns more closely with human assessment than the system that uses linear models.
Performance was improved by first translating the ASR transcripts before feeding the result to an LLM.
In some cases, a human expert was a poor judge when reassessing some of the stories; this revealed the complexity and subjectivity of scoring pre-preschoolers' oral narratives, 
emphasising the need for further refinement to address the nuanced aspects of automatically assessing young children's spoken language.
Nevertheless, we believe that our system will be usable in assisting teachers to identify cases where intervention is most required.

Future work will look at even more approaches for prompting the LLM and end-to-end training methods that go directly from speech to a score.

\vspace{3pt}
\noindent \textbf{Acknowledgements:} This work was supported by grants from the Het Jan Marais Fonds (HJMF) and Fab Inc.

\bibliographystyle{IEEEtran}
\bibliography{mybib}

\end{document}